\documentclass[11pt,letterpaper]{article}
\usepackage{naaclhlt2015}
\usepackage{times}
\usepackage{latexsym}
\usepackage{xcolor}
\usepackage{url}
\usepackage{amsmath}
\usepackage{amsfonts}
\usepackage{graphicx}
\usepackage{multirow}
\usepackage{subcaption}

\newcommand{\model}[1]{\textsc{#1}}

\title{Combining Language and Vision\\ with a Multimodal Skip-gram
  Model}

\author{Angeliki Lazaridou~~~Nghia The Pham~~~Marco Baroni\\
  Center for Mind/Brain Sciences\\
  University of Trento\\
  {\tt \{angeliki.lazaridou|thenghia.pham|marco.baroni\}@unitn.it} \\}

\begin{document}

\maketitle

\begin{abstract}
  We extend the \model{Skip-gram} model of
  \newcite{Mikolov:etal:2013c} by taking visual information into
  account. Like \model{Skip-gram}, our \emph{multimodal} models
  (\model{MMSkip-gram}) build vector-based word representations by
  learning to predict linguistic contexts in text corpora. However,
  for a restricted set of words, the models are also exposed to visual
  representations of the objects they denote (extracted from natural
  images), and must predict linguistic and visual features jointly.
  The \model{MMSkip-gram} models achieve good performance on a variety
  of semantic benchmarks. Moreover, since they propagate visual
  information to all words, we use them to improve image labeling
  and retrieval in the zero-shot setup, where the test concepts are
  never seen during model training. Finally, the \model{MMSkip-gram} models
  discover intriguing visual properties of abstract words, paving the
  way to realistic implementations of embodied theories of meaning.
\end{abstract}

\section{Introduction}
\label{sec:introduction}

Distributional semantic models (DSMs) derive vector-based
representations of meaning from patterns of word co-occurrence in
corpora. DSMs have been very effectively applied to a variety of
semantic tasks
\cite{Clark:2012b,Mikolov:etal:2013a,Turney:Pantel:2010}. However,
compared to human semantic knowledge, these purely textual models,
just like traditional symbolic AI systems
\cite{Harnad:1990,Searle:1984}, are severely impoverished, suffering
of \textit{lack of grounding} in extra-linguistic modalities
\cite{Glenberg:Robertson:2000}. This observation has led to the
development of \textit{multimodal distributional semantic models}
(MDSMs) \cite{Bruni:etal:2014,Feng:Lapata:2010,Silberer:Lapata:2014},
that enrich linguistic vectors with perceptual information, most often
in the form of visual features automatically induced from image
collections.

MDSMs outperform state-of-the-art text-based approaches, not only in
tasks that directly require access to visual knowledge
\cite{Bruni:etal:2012a}, but also on general semantic benchmarks
\cite{Bruni:etal:2014,Silberer:Lapata:2014}. However, current MDSMs
still have a number of drawbacks. First, they are generally
constructed by first separately building linguistic and visual
representations of the same concepts, and then merging them. This is
obviously very different from how humans learn about concepts, by
hearing words in a situated perceptual context. Second, MDSMs assume
that both linguistic and visual information is available for all
words, with no generalization of knowledge across modalities. Third,
because of this latter assumption of full linguistic and visual
coverage, current MDSMs, paradoxically, cannot be applied to computer
vision tasks such as image labeling or retrieval, since they do not
generalize to images or words beyond their training set.

We introduce the \emph{multimodal skip-gram} models, two new MDSMs
that address all the issues above. The models build upon the very
effective skip-gram approach of \newcite{Mikolov:etal:2013c}, that
constructs vector representations by learning, incrementally, to
predict the linguistic contexts in which target words occur in a
corpus. In our extension, for a subset of the target words, relevant
visual evidence from natural images is presented together with the
corpus contexts (just like humans hear words accompanied by concurrent
perceptual stimuli). The model must learn to predict these visual
representations jointly with the linguistic features. The joint
objective encourages the propagation of visual information to
representations of words for which no direct visual evidence was
available in training. The resulting multimodally-enhanced vectors
achieve remarkably good performance both on traditional semantic
benchmarks, and in their new application to the ``zero-shot'' image
labeling and retrieval scenario. Very interestingly, indirect visual
evidence also affects the representation of abstract words, paving the
way to ground-breaking cognitive studies and novel applications in
computer vision.

\section{Related Work}
\label{sec:related-work}

There is by now a large literature on multimodal distributional
semantic models. We focus here on a few representative
systems. \newcite{Bruni:etal:2014} propose a straightforward approach
to MDSM induction, where text- and image-based vectors for the same
words are constructed independently, and then ``mixed'' by applying
the Singular Value Decomposition to their concatenation. An
empirically superior model has been proposed by
\newcite{Silberer:Lapata:2014}, who use more advanced visual
representations relying on images annotated with high-level ``visual
attributes'', and a multimodal fusion strategy based on stacked
autoencoders. \newcite{Kiela:Bottou:2014} adopt instead a simple
concatenation strategy, but obtain empirical improvements by using
state-of-the-art convolutional neural networks to extract visual
features, and the skip-gram model for text. These and related systems
take a two-stage approach to derive multimodal spaces (unimodal
induction followed by fusion), and they are only tested on concepts
for which both textual and visual labeled training data are available
(the pioneering model of \newcite{Feng:Lapata:2010} did learn from
text and images jointly using Topic Models, but was shown to be
empirically weak by \newcite{Bruni:etal:2014}).

\newcite{Howell:etal:2005} propose an incremental multimodal model
based on simple recurrent networks \cite{Elman:1990}, focusing on
\textit{grounding propagation} from early-acquired concrete words to a
larger vocabulary.  However, they use subject-generated features as
surrogate for realistic perceptual information, and only test the
model in small-scale simulations of word
learning. \newcite{Hill:Korhonen:2014b}, whose evaluation focuses on
how perceptual information affects different word classes more or less
effectively, similarly to Howell et al., integrate perceptual
information in the form of subject-generated features and text from
image annotations into a skip-gram model. They inject perceptual
information by merging words expressing perceptual features with
corpus contexts, which amounts to linguistic-context re-weighting,
thus making it impossible to separate linguistic and perceptual
aspects of the induced representation, and to extend the model with
non-linguistic features.  We use instead authentic image analysis as
proxy to perceptual information, and we design a robust way to
incorporate it, easily extendible to other signals, such as feature
norm or brain signal vectors~\cite{Fyshe:etal:2014}.

The recent work on so-called \emph{zero-shot learning} to address the
annotation bottleneck in image labeling
\cite{Frome:etal:2013,Lazaridou:etal:2014,Socher:etal:2013a} looks at
image- and text-based vectors from a different perspective. Instead of
combining visual and linguistic information in a common space, it aims
at learning a mapping from image- to text-based vectors. The mapping,
induced from annotated data, is then used to project images of objects
that were not seen during training onto linguistic space, in order to
retrieve the nearest word vectors as labels. Multimodal word vectors
should be better-suited than purely text-based vectors for the task,
as their similarity structure should be closer to that of
images. However, traditional MDSMs cannot be used in this setting,
because they do not cover words for which no manually annotated
training images are available, thus defeating the generalizing purpose
of zero-shot learning. We will show below that our multimodal vectors,
that are not hampered by this restriction, do indeed bring a
significant improvement over purely text-based linguistic
representations in the zero-shot setup.

Multimodal language-vision spaces have also been developed with the
goal of improving caption generation/retrieval and caption-based image
retrieval
\cite{Karpathy:etal:2014,Kiros:etal:2014b,Mao:etal:2014,Socher:etal:2014}. These
methods rely on necessarily limited collections of captioned images as
sources of multimodal evidence, whereas we automatically enrich a very
large corpus with images to induce general-purpose multimodal word
representations, that could be used as input embeddings in systems
specifically tuned to caption processing. Thus, our work is
complementary to this line of research.



\section{Multimodal Skip-gram Architecture} \label{sec:model}

\subsection{Skip-gram Model} \label{sec:plain} We start by reviewing
the standard \model{Skip-gram} model of \newcite{Mikolov:etal:2013c},
in the version we use. Given a text corpus, \model{Skip-gram} aims at
inducing word representations that are good at predicting the
\emph{context} words surrounding a \emph{target} word.
Mathematically, it maximizes the objective function:\\[-2ex]
\begin{equation}
\frac{1}{T} \sum_{t=1}^{T} \left(\sum_{-c\leq j\leq c, j\ne 0} \log p(w_{t+j}|w_t)\right)
\label{eq:1}
\end{equation}
where $w_1,w_2,...,w_T$ are words in the training corpus and $c$ is
the size of the window around target $w_t$, determining the set of
context words to be predicted by the induced representation of $w_t$.
Following Mikolov et al., we implement a subsampling option randomly
discarding context words as an inverse function of their frequency,
controlled by hyperparameter $t$. The probability $p(w_{t+j}|w_t)$,
the core part of the
objective in Equation~\ref{eq:1}, is given by softmax:\\[-2ex]
\begin{equation}
p(w_{t+j}|w_t) = \frac{e^{{u'_{w_{t+j}}}^T u_{w_t}}}{\sum_{w'=1}^{W} e^{{u'_{w'}}^T u_{w_t}}}
\label{eq:skip-gram}
\end{equation}
where $u_w$ and $u_w'$ are the context and target vector
representations of word $w$ respectively, and $W$ is the size of the
vocabulary.  Due to the normalization term,
Equation~\ref{eq:skip-gram} requires $O(|W|)$ time complexity. A
considerable speedup to $O(\log|W|)$, is achieved by using the
hierarchical version of Equation~\ref{eq:skip-gram}
~\cite{Morin:Bengio:2005}, adopted here.


\begin{figure}
\includegraphics[scale=0.33]{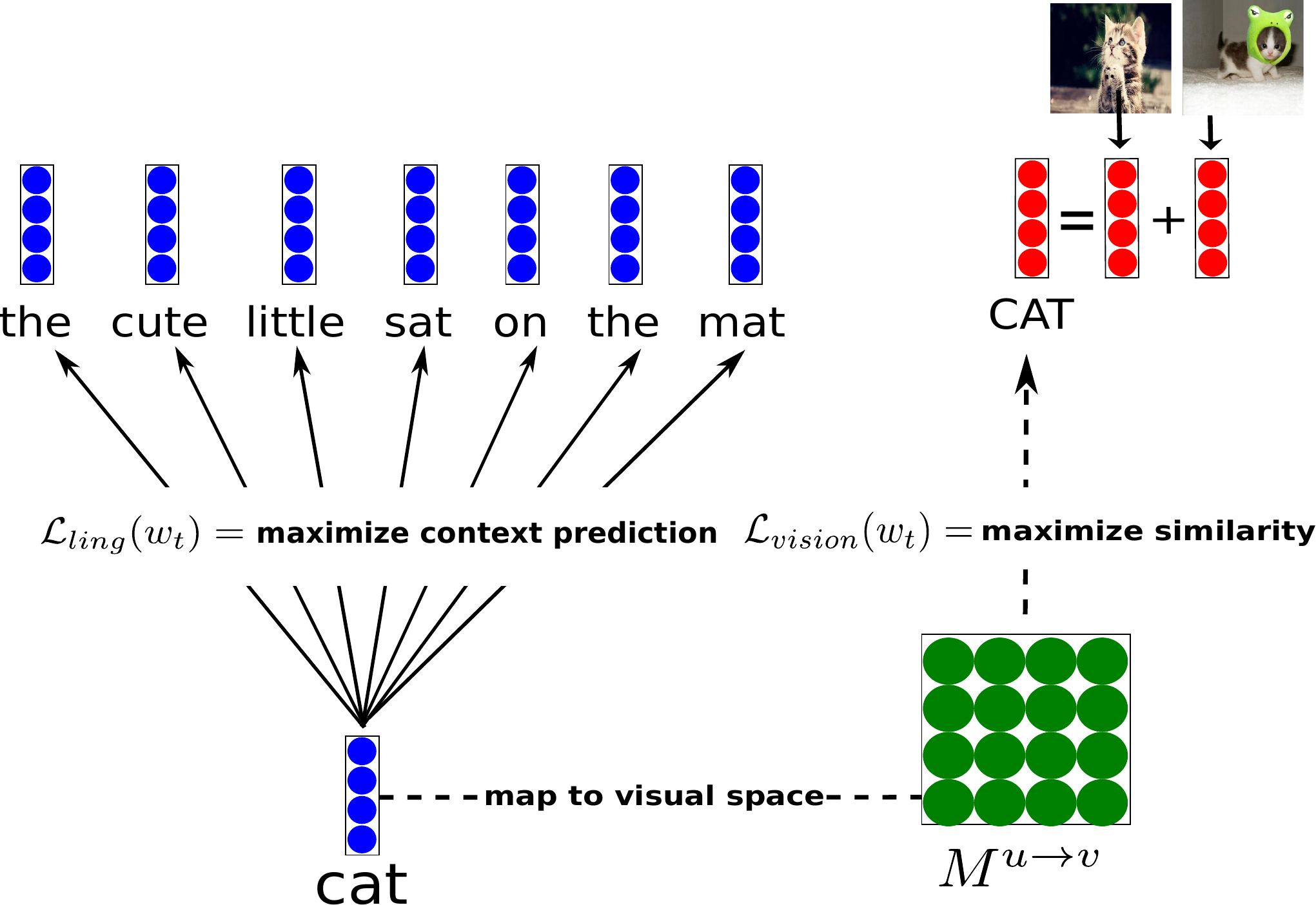}
\caption{``Cartoon'' of \model{MMSkip-gram-B}. Linguistic context
  vectors are actually associated to classes of words in a tree, not
  single words. \model{Skip-gram} is obtained by ignoring the visual
  objective, \model{MMSkip-gram-A} by fixing $M^{u\rightarrow v}$ to
  the identity matrix.}
 \label{fig:model1}
\end{figure}

\subsection{Injecting visual knowledge} 
\label{sec:multi} 
We now assume that word learning takes place in a \emph{situated}
context, in which, for a subset of the target words, the corpus
contexts are accompanied by a visual representation of the concepts
they denote (just like in a conversation, where a linguistic utterance
will often be produced in a visual scene including some of the word
referents). The visual representation is also encoded in a vector (we
describe in Section \ref{sec:experimental-setup} below how we
construct it).  We thus make the skip-gram ``multimodal'' by adding a
second, \emph{visual} term to the original \emph{linguistic}
objective, that is, we extend Equation~\ref{eq:1} as follow:\\[-2ex]
\begin{equation}
\frac{1}{T} \sum_{t=1}^{T} \left(\mathcal{L}_{ling}(w_t) + \mathcal{L}_{vision}(w_t)\right)
\label{eq:mm-objective}
\end{equation}
where $\mathcal{L}_{ling}(w_t)$ is the text-based skip-gram objective
$\sum_{-c\leq j\leq c, j\ne 0} \log p(w_{t+j}|w_t)$, whereas the
$\mathcal{L}_{vision}(w_t)$ term forces word representations to take
visual information into account. Note that if a word $w_t$ is not
associated to visual information, as is systematically the case, e.g.,
for determiners and non-imageable nouns, but also more generally for
any word for which no visual data are available,
$\mathcal{L}_{vision}(w_t)$ is set to 0.

We now propose two variants of the visual objective, resulting in two
distinguished multi-modal versions of the skip-gram model.

\subsection{Multi-modal Skip-gram Model A} 
\label{sec:model-a}
One way to force word embeddings to take visual representations into
account is to try to directly increase the similarity (expressed, for
example, by the \emph{cosine}) between linguistic and visual
representations, thus aligning the dimensions of the linguistic vector
with those of the visual one (recall that we are inducing the first,
while the second is fixed), and making the linguistic representation
of a concept ``move'' closer to its visual representation. We maximize
similarity through a max-margin framework commonly used in models
connecting language and vision
\cite{Weston:etal:2010,Frome:etal:2013}. More precisely, we formulate
the visual objective $\mathcal{L}_{vision}(w_t)$ as:\\[-2ex]
\begin{equation}
\footnotesize
    \label{eq:maxmargin}
- \sum_{w'\sim P_n(w)} \max(0,\gamma-cos(u_{w_t},v_{w_t})+cos(u_{w_t},v_{w'}))
\end{equation}
where the minus sign turns a loss into a cost, $\gamma$ is the margin,
$u_{w_t}$ is the target multimodally-enhanced word representation we
aim to learn, $v_{w_t}$ is the corresponding visual vector (fixed in
advance) and $v_{w'}$ ranges over visual representations of words
(featured in our image dictionary) randomly sampled from distribution
$P_n(w_t)$. These random visual representations act as ``negative''
samples, encouraging $u_{w_t}$ to be more similar to its own visual
representation than to that of other words. The sampling distribution
is currently set to uniform, and the number of negative samples
controlled by hyperparameter
$k$. 


\subsection{Multi-modal Skip-gram Model B} 
\label{sec:model-b}
The visual objective in \model{MMSkip-gram-A} has the drawback of
assuming a direct comparison of linguistic and visual representations,
constraining them to be of equal size. \model{MMSkip-gram-B} lifts
this constraint by including an extra layer mediating between
linguistic and visual representations (see Figure \ref{fig:model1} for
a sketch of \model{MMSkip-gram-B}).  Learning this layer is equivalent
to estimating a cross-modal mapping matrix from linguistic onto visual
representations, jointly induced with linguistic word embeddings. The
extension is straightforwardly implemented by substituting, into
Equation \ref{eq:maxmargin}, the word representation $u_{w_t}$ with
$z_{w_t}=M^{u\rightarrow v}u_{w_t}$, where $M^{u\rightarrow v}$ is the
cross-modal mapping matrix to be
induced. 
To avoid overfitting, we also add an L2 regularization term for
$M^{u\rightarrow v}$ to the overall objective
(Equation~\ref{eq:mm-objective}), with its relative importance
controlled by hyperparamer $\lambda$.


\section{Experimental Setup}
\label{sec:experimental-setup}

The parameters of all models are estimated by back-propagation of
error via stochastic gradient descent. Our text corpus is a Wikipedia
2009 dump comprising approximately 800M
tokens.\footnote{\url{http://wacky.sslmit.unibo.it}} To train the
multimodal models, we add visual information for 5,100 words that have
an entry in ImageNet \cite{Deng:etal:2009}, occur at least 500 times
in the corpus and have concreteness score $\geq$ 0.5 according to
\newcite{Turney:etal:2011}.  On average, about 5\% tokens in the text
corpus are associated to a visual representation. To construct the
visual representation of a word, we sample 100 pictures from its
ImageNet entry, and extract a 4096-dimensional vector from each
picture using the Caffe toolkit \cite{Jia:etal:2014}, together with
the pre-trained convolutional neural network of
\newcite{Krizhevsky:etal:2012}. The vector corresponds to activation
in the top (\model{fc7}) layer of the network. Finally, we average the
vectors of the 100 pictures associated to each word, deriving 5,100
aggregated visual representations.

\paragraph{Hyperparameters} For both \model{Skip-gram} and the
\model{MMSkip-gram} models, we fix hidden layer size to 300. To
facilitate comparison between \model{MMSkip-gram-A} and
\model{MMSkip-gram-B}, and since the former requires equal linguistic
and visual dimensionality, we keep the first 300 dimensions of the
visual vectors.  For the linguistic objective, we use hierarchical
softmax with a Huffman frequency-based encoding tree, setting
frequency subsampling option $t\!\!=\!\!0.001$ and window size
$c\!\!=\!\!5$, without tuning. The following hyperparameters were
tuned on the text9
corpus:\footnote{\url{http://mattmahoney.net/dc/textdata.html}}
\model{MMSkip-gram-A}: $k\!\!=\!\!20$, $\gamma\!\!=\!\!0.5$;
\model{MMSkip-gram-B}: $k\!\!=\!\!5$, $\gamma\!\!=\!\!0.5$,
$\lambda\!\!=\!\!0.0001$.


\section{Experiments}
\subsection{Approximating human judgments}

\paragraph{Benchmarks} A widely adopted way to test DSMs and their
multimodal extensions is to measure how well model-generated scores
approximate human similarity judgments about pairs of words. We put
together various benchmarks covering diverse aspects of meaning, to
gain insights on the effect of perceptual information on different
similarity facets.  Specifically, we test on general relatedness
(\textit{MEN}, \newcite{Bruni:etal:2014}, 3K pairs), e.g., pickles are
related to hamburgers, semantic ($\approx$ taxonomic) similarity
(\textit{Simlex-999}, \newcite{Hill:etal:2014b}, 1K pairs;
\textit{SemSim}, \newcite{Silberer:Lapata:2014}, 7.5K pairs), e.g.,
pickles are similar to onions, as well as visual similarity
(\textit{VisSim}, \newcite{Silberer:Lapata:2014}, same pairs as SemSim
with different human ratings), e.g., pickles look like zucchinis.

\paragraph{Alternative Multimodal Models} 
We compare our models against several recent alternatives.  We test
the vectors made available by \newcite{Kiela:Bottou:2014}. Similarly
to us, they derive textual features with the skip-gram model (from a
portion of the Wikipedia and the British National Corpus) and use
visual representations extracted from the ESP data-set
\cite{VonAhn:Dabbish:2004} through a convolutional neural network
\cite{Oquab:etal:2014}. They concatenate textual and visual features
after normalizing to unit length and centering to zero mean. We also
test the vectors that performed best in the evaluation of
\newcite{Bruni:etal:2014}, based on textual features extracted from a
3B-token corpus and SIFT-based Bag-of-Visual-Words visual features
\cite{Sivic:2003} extracted from the ESP collection. Bruni and
colleagues fuse a weighted concatenation of the two components through
SVD. %
We further re-implement both methods with our own textual and visual
embeddings as \model{Concatenation} and \model{SVD} (with target
dimensionality 300, picked without tuning).  Finally, we present for
comparison the results on \textit{SemSim} and \textit{VisSim} reported
by \newcite{Silberer:Lapata:2014}, obtained with a
stacked-autoencoders architecture run on textual features extracted
from Wikipedia with the Strudel algorithm \cite{Baroni:etal:2010}
and attribute-based visual features \cite{Fahradi:etal:2009} extracted
from ImageNet.

All benchmarks contain a fair amount of words for which we did not use
direct visual evidence. We are interested in assessing the models both
in terms of how they fuse linguistic and visual evidence when they are
both available, and for their robustness in lack of full visual
coverage. We thus evaluate them in two settings. The visual-coverage
columns of Table \ref{tab:benchmarks} (those on the right) report
results on the subsets for which all compared models have access to
direct visual information for both words. We further report results on
the full sets (``100\%'' columns of Table \ref{tab:benchmarks}) for
models that can propagate visual information and that, consequently,
can meaningfully be tested on words without direct visual
representations. %

\begin{table*}
\begin{center}
\footnotesize
\begin{tabular}{l|l l|l l|l l|l l}
\multirow{2}{*}{\textbf{Model}}  &\multicolumn{2}{c|}{\textit{MEN}}    &\multicolumn{2}{c|}{\textit{Simlex-999}} &\multicolumn{2}{c|}{\textit{SemSim}}&\multicolumn{2}{c}{\textit{VisSim}} \\
                                 &  100\%     &      42\%             &   100\%        &   29\%            &   100\%         & 85\%                    & 100\%         & 85\%  \\
\hline
\model{Kiela and Bottou}   & 	-	   &      0.74  	  & 	-  &   0.33		      &   -	      & 0.60			& -		& 0.50 \\
\model{Bruni et al.}       & 	-	   &      0.77  	  & 	-  &   0.44		      &   -	      & 0.69			& -		& 0.56 \\
\model{Silberer and Lapata}& 	-	   &      -	 	  & 	-  &  -			      &   0.70	      & -			& 0.64		& -    \\\hline
\model{CNN features}             &   -         &      0.62  	  &    -	 &   0.54			      &   -	      & 0.55			& -		& 0.56 \\
\model{Skip-gram}                &   0.70	   &      0.68  	  &    0.33  &   0.29		      &   0.62	      & 0.62			& 0.48		& 0.48 \\
\hline                                                                
\model{Concatenation}            & 	-	   &      0.74  	  & 	-  &   0.46		      &   -	      & 0.68			& -		& 0.60 \\
\model{SVD}                  &   0.61	   &      0.74  	  &    0.28  &   0.46		      &   0.65	      & 0.68			& 0.58		& 0.60 \\
\model{MMSkip-gram-A}		 &   0.75	   &      0.74  	  &    0.37  &   0.50		      &   0.72	      & 0.72			& 0.63		& 0.63 \\
\model{MMSkip-gram-B}		 &   0.74	   &      0.76  	  &    0.40  &   0.53		      &   0.66	      & 0.68			& 0.60		& 0.60 \\
\end{tabular}
\caption{Spearman correlation between model-generated similarities and
  human judgments. Right columns report correlation on visual-coverage
  subsets (percentage of original benchmark covered by subsets on first
  row of respective columns). First block reports results for
  out-of-the-box models; second block for visual and textual
  representations alone; third block for our implementation of
  multimodal models.}
\label{tab:benchmarks}
\end{center}
\end{table*}

\paragraph{Results} The state-of-the-art visual \model{CNN features}
alone perform remarkably well, outperforming the purely textual model
(\model{Skip-gram}) in two tasks, and achieving the best absolute
performance on the visual-coverage subset of Simlex-999. Regarding
multimodal \emph{fusion} (that is, focusing on the visual-coverage
subsets), both \model{MMSkip-gram} models perform very well, at the
top or just below it on all tasks, with comparable results for the two
variants. Their performance is also good on the full data sets, where
they consistently outperform \model{Skip-gram} and \model{SVD} (that
is much more strongly affected by lack of complete visual
information). They're just a few points below the state-of-the-art MEN
correlation (0.8), achieved by \newcite{Baroni:etal:2014} with a
corpus 3 larger than ours and extensive tuning. \model{MMSkip-gram-B}
is close to the state of the art for Simlex-999, reported by the
resource creators to be at 0.41~\cite{Hill:etal:2014b}. Most
impressively, \model{MMSkip-gram-A} reaches the performance level of
the \newcite{Silberer:Lapata:2014} model on their SemSim and VisSim
data sets, despite the fact that the latter has full visual-data
coverage and uses attribute-based image representations, requiring
supervised learning of attribute classifiers, that achieve performance in 
the semantic tasks comparable or higher than that of our CNN features (see Table 3 in 
\newcite{Silberer:Lapata:2014}).
Finally, if the multimodal models (unsurprisingly)
bring about a large performance gain over the purely linguistic model
on visual similarity, the improvement is consistently large also for
the other benchmarks, confirming that multimodality leads to better
semantic models in general, that can help in capturing different types
of similarity (general relatedness, strictly taxonomic, perceptual).

\begin{table*}
\begin{center}
\footnotesize
\begin{tabular}{l|c|c|c}
  \emph{Target}  &\model{Skip-gram} &\model{MMSkip-gram-A}&\model{MMSkip-gram-B}\\\hline
  donut& fridge, diner, candy&pizza, sushi, sandwich&pizza, sushi, sandwich\\
  owl& pheasant, woodpecker, squirrel&eagle, woodpecker, falcon&eagle, falcon, hawk\\\hline
  mural& sculpture, painting, portrait&painting, portrait, sculpture& painting, portrait, sculpture\\
  tobacco& coffee, cigarette, corn&cigarette, cigar, corn& cigarette, cigar, smoking\\
  depth&size, bottom, meter&sea, underwater, level&sea, size, underwater\\
  chaos& anarchy, despair, demon& demon, anarchy, destruction& demon, anarchy, shadow\\
\end{tabular}
\caption{Ordered top 3 neighbours of example words in purely textual
  and multimodal spaces. Only \emph{donut} and \emph{owl} were
  trained with direct visual information.}
\label{tab:textual-multimodal-nns}
\end{center}
\end{table*}

While we defer to further work a better understanding of the relation
between multimodal grounding and different similarity relations, Table
\ref{tab:textual-multimodal-nns} provides qualitative insights on how
injecting visual information changes the structure of semantic
space. The top \model{Skip-gram} neighbours of \emph{donuts} are
places where you might encounter them, whereas the multimodal models
relate them to other take-away food, ranking visually-similar pizzas
at the top. The \emph{owl} example shows how multimodal models pick
taxonomically closer neighbours of concrete objects, since often
closely related things also look similar \cite{Bruni:etal:2014}. In
particular, both multimodal models get rid of squirrels and offer
other birds of prey as nearest neighbours. No direct visual evidence
was used to induce the embeddings of the remaining words in the table,
that are thus influenced by vision only by propagation. The subtler
but systematic changes we observe in such cases suggest that this
indirect propagation is not only non-damaging with respect to purely
linguistic representations, but actually beneficial. For the concrete
\emph{mural} concept, both multimodal models rank paintings and
portraits above less closely related sculptures (they are not a form
of painting). For \emph{tobacco}, both models rank cigarettes and
cigar over coffee, and \model{MMSkip-gram-B} avoids the arguably less
common ``crop'' sense cued by corn. The last two examples show how the
multimodal models turn up the embodiment level in their representation
of abstract words. For \emph{depth}, their neighbours suggest a
concrete marine setup over the more abstract measurement sense picked
by the \model{MMSkip-gram} neighbours. For \emph{chaos}, they rank a
demon, that is, a concrete agent of chaos at the top, and replace the
more abstract notion of despair with equally gloomy but more imageable
shadows and destruction (more on abstract words below).\\[-3ex]


\subsection{Zero-shot image labeling and retrieval}

The multimodal representations induced by our models should be better
suited than purely text-based vectors to label or retrieve images. In
particular, given that the quantitative and qualitative results
collected so far suggest that the models propagate visual information
across words, we apply them to image labeling and retrieval in the
challenging zero-shot setup (see Section \ref{sec:related-work}
above).\footnote{We will refer here, for conciseness' sake, to
  \emph{image} labeling/retrieval, but, as our visual vectors are
  aggregated representations of images, the tasks we're modeling
  consist, more precisely, in labeling a set of pictures denoting the
  same object and retrieving the corresponding set given the name of
  the object.}

\paragraph{Setup} We take out as test set 25\% of the 5.1K words we
have visual vectors for. The multimodal models are re-trained without
visual vectors for these words, using the same hyperparameters as
above. For both tasks, the search for the correct word label/image is
conducted on the whole set of 5.1K word/visual vectors.

In the image labeling task, given a visual vector representing an
image, we map it onto word space, and label the image with the word
corresponding to the nearest vector. To perform the vision-to-language
mapping, we train a Ridge regression by 5-fold cross-validation on the
test set (for \model{Skip-gram} only, we also add the remaining 75\%
of word-image vector pairs used in estimating the multimodal models to
the Ridge training data).\footnote{We use one fold to tune Ridge
  $\lambda$, three to estimate the mapping matrix and test in the last
  fold. To enforce strict zero-shot conditions, we exclude from the
  test fold labels occurring in the LSVRC2012 set that was
  employed to train the CNN of \newcite{Krizhevsky:etal:2012}, that we
  use to extract visual features.}

In the image retrieval task, given a linguistic/multimodal vector, we
map it onto visual space, and retrieve the nearest image. For
\model{Skip-gram}, we use Ridge regression with the same training
regime as for the labeling task. For the multimodal models, since
maximizing similarity to visual representations is already part of
their training objective, we do not fit an extra mapping function. For
\model{MMSkip-gram-A}, we directly look for nearest neighbours of the
learned embeddings in visual space. For \model{MMSkip-gram-B}, we use
the $M^{u\rightarrow v}$ mapping function induced while learning word
embeddings.


\paragraph{Results} In image labeling (Table \ref{tab:labeling})
\model{Skip-gram} is outperformed by both multimodal models,
confirming that these models produce vectors that are directly
applicable to vision tasks thanks to visual propagation. 
The most interesting results however are achieved in image retrieval
(Table~\ref{tab:retrieval}), which is essentially the task the
multimodal models have been implicitly optimized for, so that they
could be applied to it without any specific training.  The strategy of
directly querying for the nearest visual vectors of the
\model{MMSkip-gram-A} word embeddings works remarkably well,
outperforming on the higher ranks \model{Skip-gram}, which requires an
ad-hoc mapping function. This suggests that the multimodal embeddings
we are inducing, while general enough to achieve good performance in
the semantic tasks discussed above, encode sufficient visual
information for direct application to image analysis tasks.  This is
especially remarkable because the word vectors we are testing were not
matched with visual representations at model training time, and are
thus multimodal only by propagation. The best performance is achieved
by \model{MMSkip-gram-B}, confirming our claim that its
$M^{u\rightarrow v}$ matrix acts as a multimodal mapping function.
\begin{table}
\begin{center}
\footnotesize
\begin{tabular}{@{}l | l| l| l | l| l}
       			&  P@1        &   P@2		&  P@10	& P@20 & P@50\\\hline
\model{Skip-gram}    	&   1.5       & 2.6 		& 14.2 	& 23.5 & 36.1\\
\model{MMSkip-gram-A}   &   2.1       & 3.7		& 16.7	& 24.6	& 37.6\\
\model{MMSkip-gram-B} 	&   2.2       & 5.1		& 20.2	& 28.5	& 43.5\\
\end{tabular}
\caption{Percentage precision@$k$ results in the zero-shot image
  labeling task.} 
\label{tab:labeling}
\end{center}
\end{table}

\begin{table}
\begin{center}
\footnotesize
\begin{tabular}{ @{}l | l| l| l | l| l}
                        &  P@1          &   P@2         &  P@10 & P@20 & P@50\\\hline
\model{Skip-gram}       &   1.9       & 3.3            & 11.5  & 18.5 & 30.4\\
\model{MMSkip-gram-A}   &   1.9       & 3.2             & 13.9  & 20.2  & 33.6\\
 \model{MMSkip-gram-B}  &   1.9       & 3.8             & 13.2  & 22.5  & 38.3\\
\end{tabular}
\caption{Percentage precision@$k$ results in the zero-shot image
  retrieval task.}
\label{tab:retrieval}
\end{center}
\end{table}


\subsection{Abstract words}


We have already seen, through the \emph{depth} and \emph{chaos}
examples of Table \ref{tab:textual-multimodal-nns}, that the indirect
influence of visual information has interesting effects on the
representation of abstract terms. The latter have received little
attention in multimodal semantics, with \newcite{Hill:Korhonen:2014b}
concluding that abstract nouns, in particular, do not benefit from
propagated perceptual information, and their representation is even
harmed when such information is forced on them (see Figure 4 of their
paper).  Still, embodied theories of cognition have provided
considerable evidence that abstract concepts are also grounded in the
senses \cite{Barsalou:2008,Lakoff:Johnson:1999}. Since the word
representations produced by \model{MMSkip-gram-A}, including those
pertaining to abstract concepts, can be directly used to search for
near images in visual space, we decided to verify, experimentally, if
these near images (of concrete things) are relevant not only for
concrete words, as expected, but also for abstract ones, as predicted
by embodied views of meaning.

More precisely, we focused on the set of 200 words that were sampled
across the USF norms concreteness spectrum by
\newcite{Kiela:etal:2014} (2 words had to be excluded for technical
reasons). This set includes not only concrete (\emph{meat}) and
abstract (\emph{thought}) nouns, but also adjectives (\emph{boring}),
verbs (\emph{teach}), and even grammatical terms (\emph{how}). Some
words in the set have relatively high concreteness ratings, but are
not particularly imageable, e.g.: \emph{hot, smell, pain, sweet}. For
each word in the set, we extracted the nearest neighbour picture of
its \model{MMSkip-gram-A} representation, and matched it with a random
picture. The pictures were selected from a set of 5,100, all labeled
with distinct words (the picture set includes, for each of the words
associated to visual information as described in Section
\ref{sec:experimental-setup}, the nearest picture to its aggregated
visual representation). Since it is much more common for concrete than
abstract words to be directly represented by an image in the picture
set, when searching for the nearest neighbour we excluded the picture
labeled with the word of interest, if present (e.g., we excluded the
picture labeled \emph{tree} when picking the nearest neighbour of the
word \emph{tree}). We ran a
CrowdFlower\footnote{\url{http://www.crowdflower.com}} survey in which
we presented each test word with the two associated images
(randomizing presentation order of nearest and random picture), and
asked subjects which of the two pictures they found more closely
related to the word. We collected minimally 20 judgments per
word. Subjects showed large agreement (median proportion of majority
choice at 90\%), confirming that they understood the task and behaved
consistently.

We quantify performance in terms of proportion of words for which the
number of votes for the nearest neighbour picture is significantly
above chance according to a two-tailed binomial test. We set
significance at $p\!\!<\!\!0.05$ after adjusting all p-values with the
Holm correction for running 198 statistical tests. The results in
Table \ref{tab:cf-survey} indicate that, in about half the cases, the
nearest picture to a word \model{MMSkip-gram-A} representation is
meaningfully related to the word. As expected, this is more often the
case for concrete than abstract words. Still, we also observe a
significant preference for the model-predicted nearest picture for
about one fourth of the abstract terms.  Whether a word was exposed to
direct visual evidence during training is of course making a big
difference, and this factor interacts with concreteness, as only two
abstract words were matched with images during training.\footnote{In
  both cases, the images actually depict concrete senses of the words:
  a memory board for \emph{memory} and a stop sign for \emph{stop}.}
When we limit evaluation to word representations that were not exposed
to pictures during training, the difference between concrete and
abstract terms, while still large, becomes less dramatic than if all
words are considered.

\begin{table}[tb]
  \begin{small}
  \begin{tabular}{l|cc|cc}
    & \emph{global}      & \emph{$|$words$|$}  & \emph{unseen} & \emph{$|$words$|$}\\
    \hline
    all      & 48\%          & 198   &30\%      & 127\\
    concrete & 73\%        & 99    &53\%     & 30 \\
    abstract & 23\%        & 99    &23\%      & 97 \\
  \end{tabular}
  \end{small}
  \caption{Subjects' preference for nearest visual neighbour of words in Kiela et al.~(2014) vs.~random pictures. Figure of merit is percentage proportion of significant results in favor of nearest neighbour  across words. Results are reported for the whole set, as well as  for words above (\emph{concrete}) and below (\emph{abstract}) the concreteness rating median. The \emph{unseen} column reports results when words exposed to direct visual evidence during training are discarded. The \emph{words} columns report set cardinality.}
  \label{tab:cf-survey}
\end{table}

\begin{figure}[tb]
\begin{center}
 \includegraphics[scale=0.6]{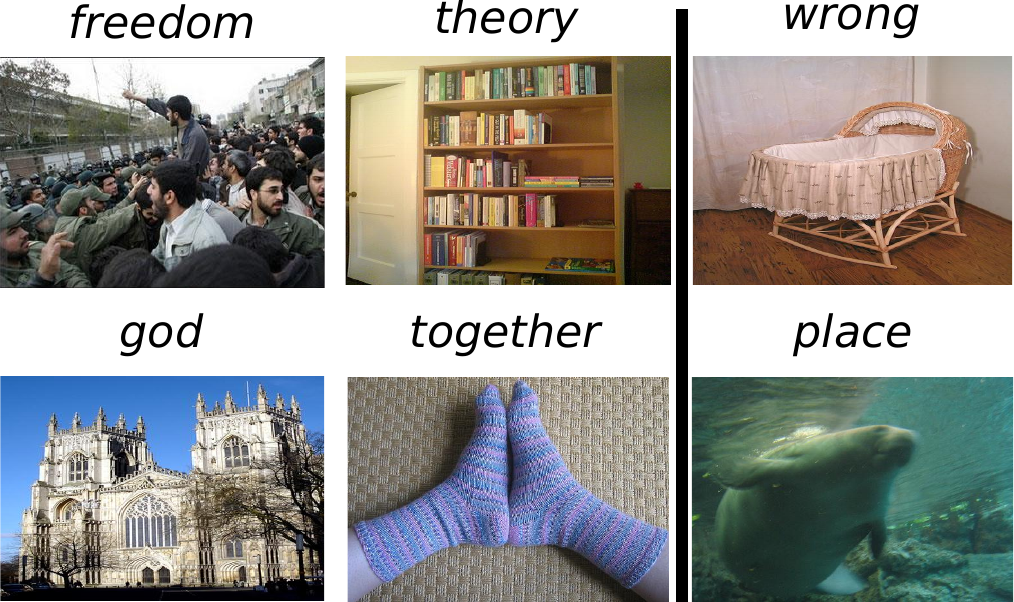}
 \caption{Examples of nearest visual neighbours of some abstract
   words: on the left, cases where subjects preferred the neighbour to
   the random foil; on the right, cases where they did not.}
\label{tab:examples}
\end{center}
\end{figure}

Figure~\ref{tab:examples} shows four cases in which subjects expressed
a strong preference for the nearest visual neighbour of a
word. \emph{Freedom, god} and \emph{theory} are strikingly in
agreement with the view, from embodied theories, that abstract words
are grounded in relevant concrete scenes and situations. The
\emph{together} example illustrates how visual data might ground
abstract notions in surprising ways. For all these cases, we can
borrow what \newcite{Howell:etal:2005} say about visual propagation to
abstract words (p.~260):
\begin{quote}
\footnotesize
Intuitively, this is something like trying to
explain an abstract concept like \emph{love} to a child by using
concrete examples of scenes or situations that are associated with
love.  The abstract concept is never fully grounded in external
reality, but it does inherit some meaning from the more concrete
concepts to which it is related.
\end{quote}

Of course, not all examples are good: the last column of
Figure~\ref{tab:examples} shows cases with no obvious relation
between words and visual neighbours (subjects preferred the
random images by a large margin).

\begin{figure}[t]
\begin{subfigure}[b]{0.3\textwidth}
\includegraphics[scale=0.29]{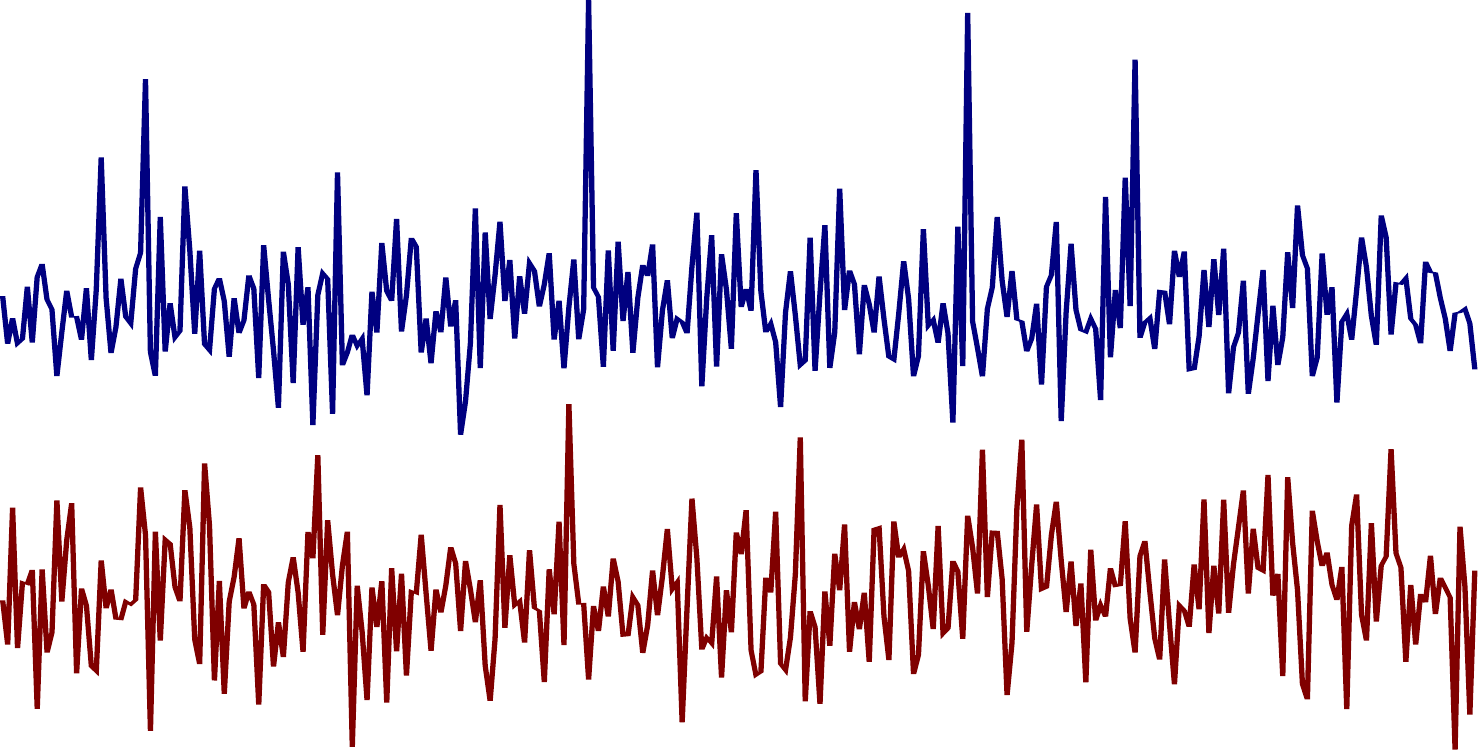}
\caption{}
\end{subfigure}
\kern-1.4em
\begin{subfigure}[b]{0.2\textwidth}
\footnotesize
\begin{tabular}[b]{@{}l@{} | r@{} }
                \textbf{Model}          &   $\rho$ \\\hline
                \model{Word frequency}  &   0.22       \\
                \model{Kiela et al.} & -0.65    \\\hline\hline
		\model{Skip-gram} & 0.05\\
		\model{MMSkip-gram-B} & 0.04\\\hline
                \model{MMSkip-gram-A} &   \textbf{-0.75}       \\
                \model{MMSkip-gram-B*} & -0.71    \\
                \end{tabular}
\caption{}
\end{subfigure}
\caption{(a) Distribution of \model{MMSkip-gram-A} vector
  activation for \emph{meat} (blue) and \emph{hope} (red). (b)
  Spearman $\rho$ between concreteness and various measures on the
  \protect\newcite{Kiela:etal:2014} set.}
\label{fig:vectors}
\end{figure}

The multimodal vectors we induce also display an interesting intrinsic
property related to the hypothesis that grounded representations of
abstract words are more complex than for concrete ones, since abstract
concepts relate to varied and composite situations
\cite{Barsalou:WiemerHastings:2005}. %
A natural corollary of this idea is that visually-grounded
representations of abstract concepts should be more diverse: If you
think of dogs, very similar images of specific dogs will come to
mind. You can also imagine the abstract notion of freedom, but the
nature of the related imagery will be much more varied. Recently,
\newcite{Kiela:etal:2014} have proposed to measure abstractness by
exploiting this very same intuition. However, they rely on manual
annotation of pictures via Google Images and define an \emph{ad-hoc}
measure of \emph{image dispersion}. We conjecture that the
representations naturally induced by our models display a similar
property.  In particular, the \emph{entropy} of our multimodal
vectors, being an expression of how varied the information they encode
is, should correlate with the degree of abstractness of the
corresponding words. As Figure \ref{fig:vectors}(a) shows, there is
indeed a difference in entropy between the
most concrete (\textit{meat}) and most abstract (\textit{hope}) words
in the Kiela et al.~set. 


To test the hypothesis quantitatively, we measure the correlation of
entropy and concreteness on the 200 words in the
\newcite{Kiela:etal:2014} set.\footnote{Since the vector dimensions
  range over the real number line, we calculate entropy on vectors
  that are unit-normed after adding a small constant insuring all
  values are positive.}  Figure~\ref{fig:vectors}(b) shows that the
entropies of both the \model{MMSkip-gram-A} representations and those
generated by mapping \model{MMSkip-gram-B} vectors onto visual space
(\model{MMSkip-gram-B*}) achieve very high correlation (but,
interestingly, not \model{MMSkip-gram-B}). This is further evidence that
multimodal learning is grounding the representations of both concrete
and abstract words in meaningful ways.

\section{Conclusion}
\label{sec:conc}

We introduced two multimodal extensions of
\model{Skip-gram}. \model{MMSkip-gram-A} is trained by directly
optimizing the similarity of words with their visual representations,
thus forcing maximum interaction between the two
modalities. \model{MMSkip-gram-B} includes an extra mediating layer,
acting as a cross-modal mapping component. The ability of the models
to \textit{integrate} and \textit{propagate} visual information
resulted in word representations that performed well in both semantic
and vision tasks, and could be used as input in systems benefiting
from prior visual knowledge (e.g., caption generation). Our results
with abstract words suggest the models might also help in tasks such
as metaphor detection, or even retrieving/generating pictures of
abstract concepts. Their incremental nature makes them well-suited for
cognitive simulations of grounded language acquisition, an avenue of
research we plan to explore further.

\section*{Acknowledgments}

We thank Adam Liska, Tomas Mikolov, the reviewers and the NIPS 2014
Learning Semantics audience. We were supported by ERC 2011 Starting
Independent Research Grant n.~283554 (COMPOSES).

\bibliography{../..//marco.bib,../../angeliki.bib,../../elia.bib}
\bibliographystyle{naaclhlt2015}
\end{document}